# Mitigating Multi-Stage Cascading Failure by Reinforcement Learning

Yongli Zhu[1], Chengxi Liu[2]
[1]GEIRI North America, San Jose, USA (yongli.zhu@geirina.net)
[2]Department of Energy Technology, Aalborg University, Aalborg, Denmark (cli@et.aau.dk)

*Abstract*—**This paper proposes a cascading failure mitigation strategy based on Reinforcement Learning (RL) method. Firstly, the principles of RL are introduced. Then, the Multi-Stage Cascading Failure (MSCF) problem is presented and its challenges are investigated. The problem is then tackled by the RL based on DC-OPF (Optimal Power Flow). Designs of the key elements of the RL framework (rewards, states, etc.) are also discussed in detail. Experiments on the IEEE 118-bus system by both shallow and deep neural networks demonstrate promising results in terms of reduced system collapse rates.**

*Index Terms*—**cascading failure, convolutional neural network, deep learning, optimal power flow, reinforcement learning.**

## I. INTRODUCTION

The fast development of the power grid is accompanied by various risks, among which the cascading failure is one of the most challenging issues necessary to be addressed. Previous studies focused on how to identify the failure mechanism of critical components and mitigate future cascading failures.

The cascading failure mitigation can be regarded as a stochastic dynamic programming problem with unknown information about the risk of failures. To tackle this challenge, there are basically two steps. The first step is to describe the mechanism of cascading failures by mathematical models, e.g. Stochastic Process [1], Complex Systems [2]. Although such mathematical models are based on strict theoretical foundations, they can become inaccurate due to parameter unavailability, oversimplified assumptions, etc. The second step is to find an effective mitigation strategy to reduce the risk of cascading failures. Plenty of researches try to solve this problem based on either mathematical programming methods or heuristic methods. For example, the bi-level programming is used to mitigate the cascading failures when energy storages exist [3]. In [4], a percolation theory-based algorithm is employed for mitigating cascading failures, by using UPFC (Unified Power Flow Controller) to redistribute the system power flow more evenly.

Meanwhile, some emerging artificial intelligence technologies such as reinforcement learning (RL) and deep learning (DL), have nourished both the fields of the power system and control theory [5]. In [6], the RL method is used for reactive power control. In [7], voltage restoration for an islanded microgrid is achieved via a distributed RL method. In [8], an application for disturbance classification is proposed based on image embedding and convolutional neural network (CNN). In [9], deep learning is applied in power consumption forecasting. However, the application of RL or DL for cascading failure study has not been reported. The main contributions of this paper are in two aspects:

1) Propose and formulate the concept of the *Multi-Stage Cascading Failure* (MSCF) problem.
2) Present a systematic reinforcement learning framework to fulfill the mitigation control for the MSCF problem.

The remaining parts of this paper are organized as follows. Section II introduces the principles of deep reinforcement learning, including two basic types of learning strategies. Section III investigates and proposes an RL-based control framework on the mitigation of cascading failures. Section IV presents a case study and result analysis. Finally, conclusions and future directions are given in Section V.

## II. PRINCIPLES OF DEEP REINFORCEMENT LEARNING

### A. Reinforcement learning

The RL is essentially a dynamic programming method that interacts with unknown "*environment*", as shown in Fig. 1.

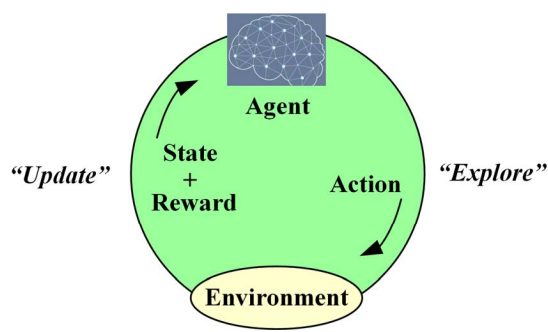

Figure 1. The basic concept of Reinforcement Learning.

Basically, there are two categories of RL methods: model-based and model-free, depending on whether the model structure information is available or not.

This work is supported by State Grid Corporation Project SG5455HJ180021

The model-based RL requires *a-prior* knowledge of system state-action relationship. For example, a robot finds the path in a labyrinth-like space with given probability distribution of each moving step towards four directions, $p(a_t$ = move north $|s_t) = 0.25$, $p(a_t$ = move south $|s_t)= 0.25$, $p(a_t$ = move east $|s_t)=0.25$, $p(a_t$ = move west $|s_t)=0.25$, where $a_t \in A=$\{move north, move south, move east, move west\} and $s_t \in S=\{(i,j) \mid i,j$ are the *x-y* coordinates of any *reachable* position in the 2D- labyrinth\}. *A* and *S* stand for the sets of action space and state space respectively.

The model-free RL means no *a-prior* information of system models is available. Take the above robot path-finding problem for example. No probability distribution of each moving-step towards the four directions is given. For most control problems in the real world, the model structure is not well known, either due to the unavailability of model/parameter information, or the high nonlinearity of the model structure. Therefore, in this paper, the model-free RL is utilized. Two mainstream approaches of model-free RL are:

*1) Value-based RL*

The value-based RL is indeed an indirect control approach, i.e. firstly a "value" $Q(s,a)$ is constructed concretely (e.g. a 2D look-up table of state-action pairs for discrete cases; or approximated by abstract means, e.g. by high-order nonlinear functions (parametric model), or by neural networks (nonparametric model). In this paper, the neural network is used. The main drawback of value-based RL is that its action space is discrete. Two representatives of the value-based RL methods, i.e. *SARSA* and *Q*-learning are adopted in this paper and are introduced in the following sub-sections.

*2) Policy-based RL*

The idea of policy-based RL is to model the policy (i.e. the probabilistic distribution of *action*) directly by neural network and train such model by RL. It can handle problems with continuous action space. This type of RL method is out of the scope of this paper. More details about this method can be found in [5].

B. *On Policy Temporal Difference (SARSA)*

The pseudo-code for *SARSA* method is shown as follows:

TABLE I. SARSA METHOD

| **SARSA (on-policy TD) method to estimate the *Q*-value** |
|---|
| 0  Initialize $Q(s, a)$ randomly, $\forall s \in S, a \in A(s)$ |
| 1  For each episode: |
| 2    Pick an initial *state s* randomly |
| 3    Pick an *action a* for *s* by *policy* derived from $Q$ ($\varepsilon$-greedy) |
| 4    For each step in current episode: |
| 5      Execute *action a*, observe reward *r* and next *state s'* |
| 6      Pick *action a'* based on *s'* by the **same** *policy* from $Q$ |
| 7      Smooth: $Q(s, a) \leftarrow (1-\alpha) Q(s, a) + \alpha [r + \gamma Q(s', a')]$ |
| 8      $s \leftarrow s', a \leftarrow a'$ |
| 9    Until $s \in S_{terminal}$ |

C. *Off Policy Temporal Difference (Q-learning)*

The pseudo-code of *Q*-learning is shown as follows:

TABLE II. Q-LEARNING METHOD

| **Q-learning (off-policy TD) method to estimate the *Q*-value** |
|---|
| 0  Initialize $Q(s, a)$ randomly, $\forall s \in S, a \in A(s)$ |
| 1  For each episode: |
| 2    Pick an initial *state s* randomly |
| 3    For each step in current episode: |
| 4      Pick an *action a* for *s* by *policy* derived from $Q$ ($\varepsilon$-greedy) |
| 5      Execute *action a*, observe reward *r* and next *state s'* |
| 6      Pick *action a'* based on *s'* by *greedy policy* derived from $Q$ |
| 7      $Q(s, a) \leftarrow (1-\alpha) Q(s, a) + \alpha [r + \gamma \max_{a'} Q(s', a')]$ (greedy) |
| 8      $s \leftarrow s'$ |
| 9    Until $s \in S_{terminal}$ |

In the above methods, $\alpha$ is the smoothing constant; $\gamma$ is the "discount factor" used for mapping the "future reward" onto the "current reward". The "$\varepsilon$-greedy" policy is given by (1) and $\varepsilon$ is a small number (e.g. 1e-4).

$$\pi(a \mid s) = \begin{cases} 1 - \varepsilon + \dfrac{\varepsilon}{|A(s)|}, & \text{if } a = \arg\max_a Q(s, a) \\ \dfrac{\varepsilon}{|A(s)|}, & \text{otherwise} \end{cases} \quad (1)$$

The above procedures can be interpreted as: in every training cycle ("*episode*"), starting from a random initial state to "play the game", and generate current *action* and next (*state, action*) pair based on certain *policy* derived indirectly from the *Q*-function. Then, the process is repeated until the "game" reaches certain terminal states. The main difference between *SARSA* and *Q-learning* is that *SARSA* method uses the *same* policy to generate *action* during the "explore phase" and "update phase"; while in *Q*-learning, different policies are adopted, i.e. "$\varepsilon$-greedy" policy is used in the *exploration* phase and "greedy" policy (seeking maximum $Q$ value) is used in the *updating* phase (for neural network based *Q*-function, the *updating* phase actually adjusts the network weights).

D. *Deep Learning*

The outstanding achievement of deep learning in image recognition area inspires researchers to apply deep neural network in RL for feature extractions. To this end, the *state* of a "game" is usually converted to a 2D image as the network input. For example, the authors of DQN (Deep *Q*-Network) [10] use screenshots of a video game as the training input and the trained networks outperform human-beings in playing that game, and thereafter the Deep Reinforcement Learning (DRL) concept is established. In this paper, a DRL approach, similar to the original DQN but with a simplified training procedure is adopted in the mitigation of cascading failures.

III. MULTI-STAGE CASCADING FAILURE CONTROL

A. *Multi-Stage Cascading Failure (MSCF) Problem*

Firstly, the following definitions are given:

*Generation*: one "event" of the cascading failures within the one stage, e.g. a line tripping.

*Stage*: after an attack (e.g. one line is broken by a natural disaster), the grid *evolves* with a series of potential *generations* (e.g. line tripping events if the line thermal limits are reached). Finally, the power system will either reach a new equilibrium point if it exists, i.e. the AC power flow (ACPF) converges and all the branch flows are within secure limits, or the system collapses.

In conventional cascading failure analysis, only one *stage* is considered, However, in certain situations, succeeding *stages* might follow shortly. For example, a wind storm results in one *generation*, in which certain lines are lost and the system reaches a new steady state. Then, shortly, a new *stage* is invoked by tripping an important line due to the misoperation of human-operator or relay protection. As an example, Table III and IV list the simulation results of the IEEE 118 system for a two-stage MSCF problem in two independent episodes.

A naïve way to handle such complicated multi-stage situations is to treat each of them independently by existing control method. However, such kind of approach may not work well due to the overlook of the correlations between any two consecutive *stages*. Thus, the MSCF problem should be considered from a *holistic* perspective.

TABLE III. RESULT OF EPISODE-1

| *Stage*-1 | | ACPF converge | Over limit Lines |
|---|---|---|---|
| | *Generation*-1 | Yes | 0 |
| *Stage*-2 | | ACPF converge | Over limit Lines |
| | *Generation*-2 | Yes | 0 |
| **Result** | Win | | |

TABLE IV. RESULT OF EPISODE-2

| *Stage*-1 | | ACPF converge | Over limit Lines |
|---|---|---|---|
| | *Generation*-1 | Yes | 2 |
| | *Generation*-2 | Yes | 0 |
| *Stage*-2 | | ACPF converge | Over limit Lines |
| | *Generation*-1 | Yes | 4 |
| | *Generation*-2 | Yes | 2 |
| | *Generation*-3 | Yes | 2 |
| | *Generation*-4 | Yes | 3 |
| | *Generation*-5 | Yes | 10 |
| | *Generation*-6 | Yes | 20 |
| | *Generation*-7 | No | -- |
| **Result** | Lose | | |

### B. Mimicking the corrective controls by DCOPF

The following DC optimal power flow (DC-OPF) is adopted to mimic the operator's corrective control measures when a cascading failure event happens. i.e. re-dispatch the generators and do load shedding (when necessary).

$$\min \sum_{i \in G} c_i p_i + \sum_{j \in D} d_j (p_j - P_{dj})$$

$$s.t. \quad \mathbf{F} = \mathbf{A}\mathbf{p}$$

$$\sum_{k=1}^{n} p_k = 0 \quad (2)$$

$$P_{dj} \leq p_j \leq 0, \ j \in D$$

$$P_{gi}^{\min} \leq p_i \leq P_{gi}^{\max}, \ i \in G$$

$$-F_l^{\max} \leq F_l \leq F_l^{\max}, \ l \in L$$

where $n$ is the total bus number. $G$, $D$, $L$ are respectively the generator set, load set and branch set; $\mathbf{F}=[F_l]$ ($l \in L$) represents the branch flow; $p_i$ ($i \in G$) is the generation dispatch for the *i*-th generator; $p_j$ ($j \in D$) is the load dispatch for the *j*-th load; $\mathbf{p}=[p_k]$, $k=\{1…n\}$ represents the (net) nodal power injections. $\mathbf{A}$ is a constant matrix to associate the (net) nodal power injections with the branch flows. $P_{dj}$ is the normal demand value for the *j*-th load; $c_i$ is the given cost coefficient of generation; $d_i$ is the given cost coefficient of load shedding. $p_i$ ($i \in G$) and $p_j$ ($j \in D$) are the decision variables for generators and load respectively.

### C. Apply RL for MSCF problem

To apply RL into a specific power system problem, the first step is to map the physical quantities to corresponding concepts of RL theory, i.e. *reward*, *state*, and *action*.

1) Reward design (of each *Stage*)
   - −Total generation cost (i.e. minus the objective value of DC-OPF) (if converge);
   - −1000 (minus one thousand if DC-OPF or AC-PF diverge); +1000 (plus one thousand if system finally reaches a *new* steady state at the last *stage*. These values are determined by trial-and-error test.

2) Action design

In the OPF, if the line flow limit is too low, the OPF might not converge due to the narrow feasible region. On the contrary, if the line flow limit is too high, the feasible region also becomes large. However, the obtained optimal solution might lead to an operation point with *tighter* power flow status on each branch, which may result in cascading failures at the next *stage* of the MSCF problem.

Thus, the "branch flow limit" $F_l^{max}$ in the above DC-OPF formulation is adopted as the *action* in the RL learning framework.

3) State design:

Several quantities of each bus and the power flow of each branch (i.e. lines and transformers) are chosen and packed as the *state* in the RL learning framework, i.e.

state=[*branch_loading_status*, $V_1$, $\theta_1$, $P_1$, $Q_1$,…,$V_n$, $\theta_n$, $P_n$, $Q_n$],

where, *branch_loading_status* are the percentage values calculated by dividing each branch flow by its loading limit for all the branches; $V_i$, $\theta_i$, $P_i$, $Q_i$ ($i=1…n$) are respectively the voltage magnitude, voltage angle, active power injection and reactive power injection of each bus.

*4) Environment*

In this study, the learning environment in the RL framework is just the power grid itself. Thus, a co-simulation platform based on the combination of DIgSILENT and MATLAB is implemented, where the commercial power system tool DIgSILENT is mainly used as the simulation engine (*environment*) to provide all needed information (*states* and *rewards*) to the RL network for learning and control. Besides, the concept of *step* within one independent *episode* corresponds to one *stage* in the MSCF problem.

Finally, the overall workflow of the grid simulation part for MSCF study is shown in Fig. 2.

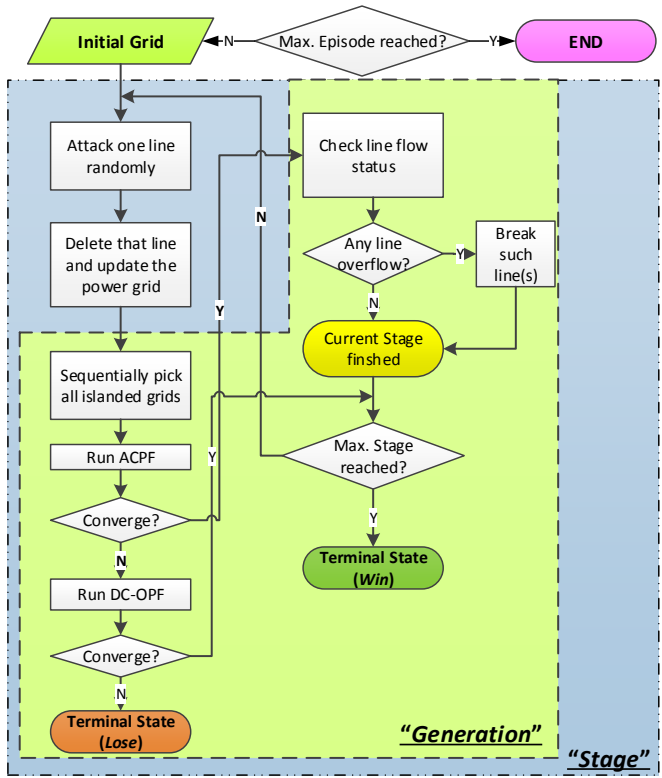

Figure 2. The overall workflow of grid simulation for MSCF study.

## IV. CASE STUDY

In this section, a modified IEEE 118-bus system is adopted as the testbed for the proposed MSCF mitigation strategy. The maximum stage number is set to 3. It contains 137 buses and 177 lines (parallel lines included), 19 generators, 34 synchronous condensers, 28 transformers and 91 loads. The system topology is shown in Fig. 3.

### A. Shallow Neural Network

The parameters set for the shallow (conventional) neural network is shown as follows, where one hidden layer with 10 neuron units is added. Its input is a 1-D vector with 753 (=137×4+177+28) elements; the output is the *action* in the RL framework (i.e. the line flow limit, c.f. Section III).

Since both the hidden-layer dimension and output-layer dimension of the shallow network is one, the *SARSA* (On-policy TD) method is applied to it. During the training, the *action* is bounded by the range [0.80, 1.25]. The network is shown in Fig. 4.

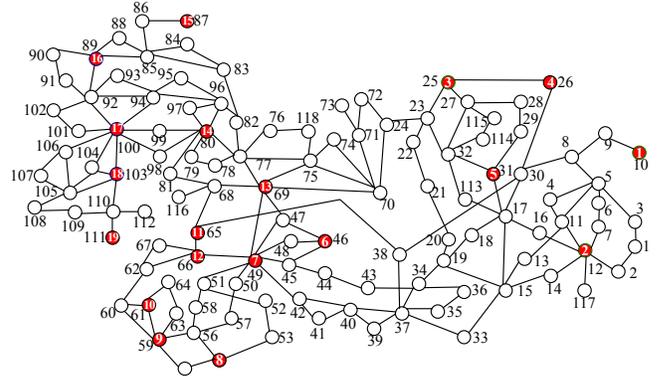

Figure 3. The topology of IEEE 118-bus system.

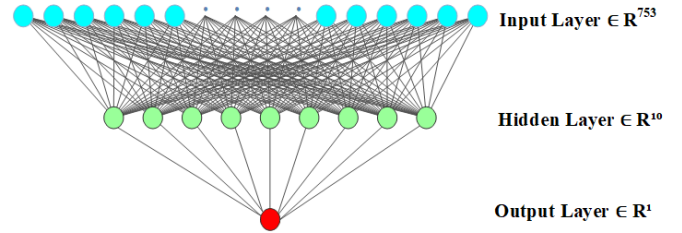

Figure 4. The shallow neural network structure used in RL.

### B. Deep Neural Network

*1) Feature engineering*

To create an image-like input to utilize the CNN, the original input (length = 753) is extended to the length of 784 = 28×28 by appending extra zeros. One example of such encoded "images" for the input *state* is shown in Fig. 5.

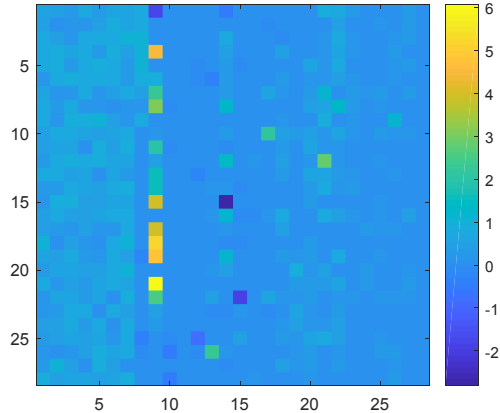

Figure 5. The image feature of one example input (*state*).

*2) Network structure*

Typically, deeper network and more layers might lead to overfitting in practice. Thus, the network structure used in this paper is shown in Fig. 6.

The *Q*-learning (Off-policy TD) method is applied to it. The output of the 2nd-last layer (dimension 1×10) will be used in both "$\varepsilon$-greedy" policy and "greedy" policy. The last-layer output (dimension 1×1) will be finally used to update the *Q*-network parameters. The candidate set of *action* is [0.8, 0.85, 0.9, 0.95, 1.0, 1.05, 1.1, 1.15, 1.20, 1.25]$\in \mathbf{R}^{10}$.

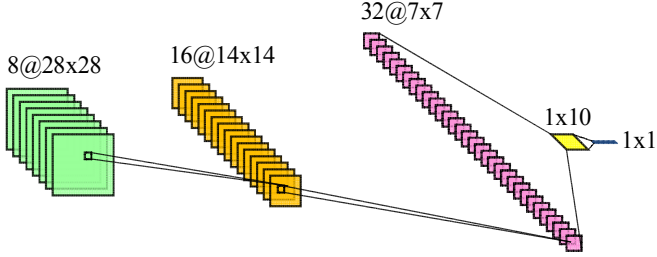

Figure 6. The network structure for Deep Reinforcement Learning.

### C. Results by RL

The parameter settings regarding the simulation and RL framework are shown in Table. V. The learning results are shown in Table. VI. The plot of moving average *reward* (window size = 50) for deep network case is shown in Fig. 7.

It can be observed that both RL and DRL have achieved satisfactory results in terms of higher *winning* rates, (equivalently, i.e. lower cascading risks). In both cases, the average *reward* per episode is more than half of the maximum possible reward (1000), which shows a positive learning ability of the RL agent in mitigating cascading failures.

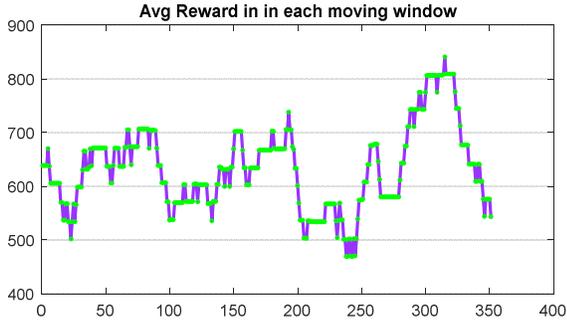

Figure 7. The moving-average reward by the DRL.

TABLE V. PARAMETER SETTINGS FOR RL

|  | Shallow Network | Deep Network |
|---|---|---|
| Episodes | 400 | 400 |
| Max Stage | 3 | 3 |
| $\gamma$ | 0.7 | 0.7 |
| $\varepsilon$ | 1e-4 | 1e-4 |

TABLE VI. LEARNING RESULTS

|  | Shallow Network | Deep Network |
|---|---|---|
| *Win* rate (moving_avg.) | 74.50% | 77.25% |
| Total *Win* times | 298 | 309 |
| Avg. *Reward* | 579.32 | 626.35 |

## V. CONCLUSIONS

In this paper, a reinforcement learning based mitigation strategy for Multi-Stage Cascading Failure problem is invented. The trained network (agent) works well on the IEEE 118-bus system by both shallow and deep neural networks. Future work includes the investigation of the effects of hyper-parameters (layer numbers, learning rate, discount factor, etc.) of the neural networks on the mitigation performance. In addition, other advanced RL techniques, e.g. policy gradient, is also left to be explored for mitigating cascading failures.